\pdfoutput=1

\documentclass[11pt]{article}

\usepackage[preprint]{acl}

\usepackage{times}
\usepackage{latexsym}

\usepackage[T1]{fontenc}

\usepackage[utf8]{inputenc}

\usepackage{microtype}

\usepackage{inconsolata}

\usepackage{graphicx}
\usepackage{wrapfig}
\usepackage{float}
\usepackage{booktabs}
\usepackage{multirow}
\usepackage{array}
\usepackage{CJKutf8}
\usepackage{longtable}
\usepackage{cite}
\usepackage{hyperref}
\usepackage[perpage]{footmisc}

\title{CFunModel: A "Funny" Language Model Capable of Chinese Humor Generation and Processing}
\author{Zhenghan Yu, Xinyu Hu, Xiaojun Wan\\
Wangxuan Institute of Computer Technology, Peking University\\
\texttt{zhenghanyu@stu.pku.edu.cn} \\
\texttt{\{huxinyu,wanxiaojun\}@pku.edu.cn}
}

\begin{document}
\begin{CJK}{UTF8}{gbsn}
\maketitle  
\begin{abstract}
Humor plays a significant role in daily language communication. With the rapid development of large language models (LLMs), natural language processing has made significant strides in understanding and generating various genres of texts. However, most LLMs exhibit poor performance in generating and processing Chinese humor. In this study, we introduce a comprehensive Chinese humor-related dataset, the \textbf{Chinese Fun Set (CFunSet)}. This dataset aggregates existing Chinese humor datasets and includes over 20,000 jokes collected from Tieba—JokeBar, a Chinese online platform known for joke sharing. The resulting corpus comprises more than 160,000 entries. Leveraging CFunSet, we developed the \textbf{Chinese Fun Model (CFunModel)}, the first large language model designed to handle various Chinese humor-related tasks including Crosstalk Response Selection, Humor Recognition, Joke Generation, etc. Experimental results demonstrate that CFunModel outperforms popular large language models in these tasks. Our CFunSet is available at \url{https://huggingface.co/datasets/ZhenghanYU/CFunSet} and CFunModel is available at \url{https://huggingface.co/ZhenghanYU/CFunModel}.
A demostration video of our work is available at \url{https://youtu.be/MOsISOJ66Ms}.
\end{abstract}

\begin{figure}[ht]
    \centering
    \includegraphics[width=1.0\linewidth]{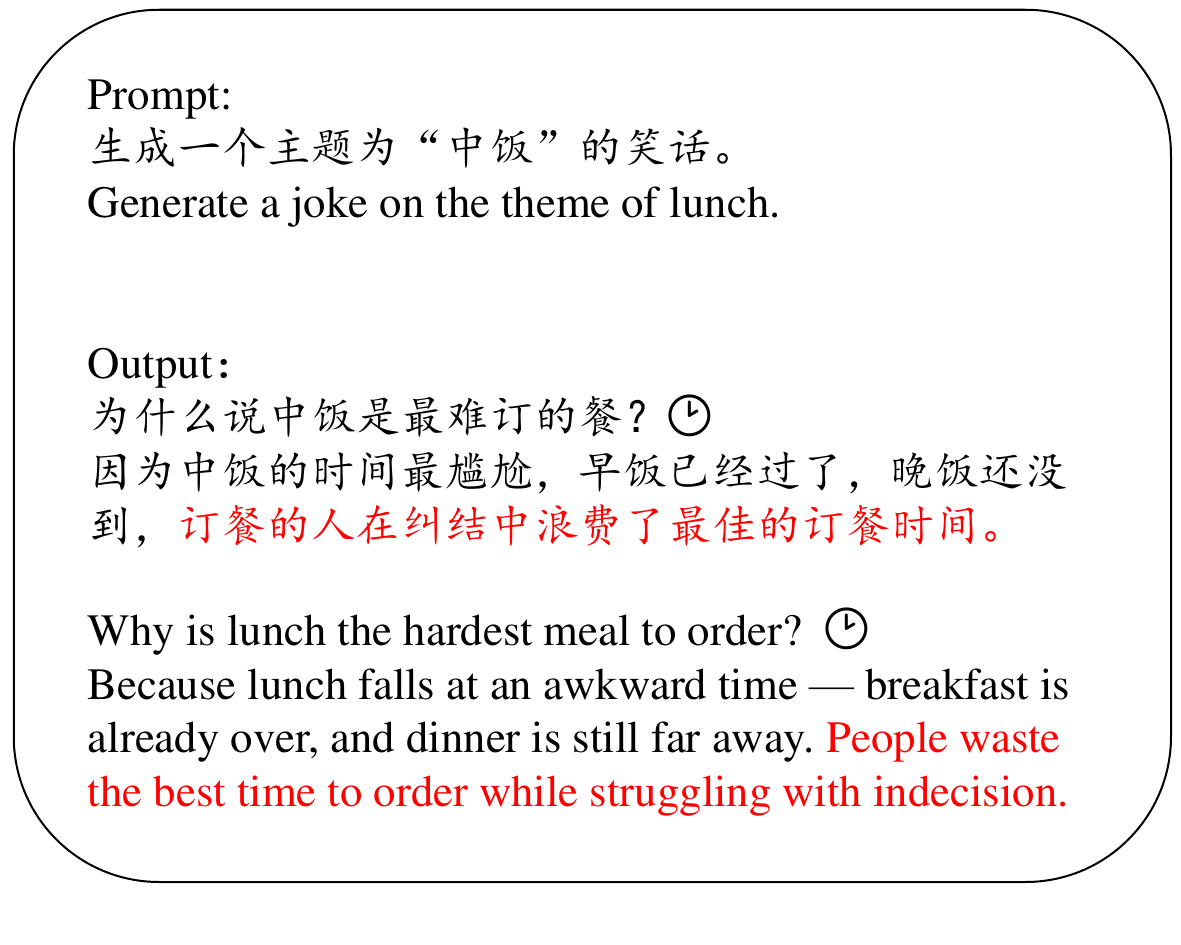}
    \caption{Examples generated by Qwen2.5-7B-Instruct on the themes of lunch. This text is intended as a joke, but it falls short in delivering effective humor. The premise sets up a relatable situation where lunch occurs between breakfast and dinner, but the sentence highlighted in red lacks the punch typically needed for a joke, only conveying a sense of mild frustration or regret.}
    \label{fig: badjoke}
\end{figure}

\section{Introduction}
Humor is an essential aspect of daily language communication, facilitating emotional expression and fostering social connections. However, developing a sense of humor requires a deep understanding of cultural contexts and extensive knowledge. While humor is captivating and engaging, it is also inherently complex and demands creativity.

With advancements in large language models (LLMs), these models have demonstrated proficiency in various natural language processing (NLP) tasks, including text generation and emotion classification. Nevertheless, LLMs encounter significant challenges in comprehending and generating humor. This difficulty arises primarily from two factors. First, humor often relies on intricate cultural backgrounds and specialized knowledge, making it difficult for LLMs to accurately interpret and reproduce. Second, the resources for Chinese humor-related datasets are also not abundant. The existing datasets predominantly focus on a single type of humor-related task and lack diversity. Without access to a sufficiently large and diverse training corpus, LLMs struggle to handle humor-related tasks effectively. Figure \ref{fig: badjoke} presents two examples generated by Qwen2.5-7B-Instruct\footnote{\url{https://qwenlm.github.io/blog/qwen2.5/}}, which exhibit a degree of coherence but lack humor.

Although humor-related tasks remain challenging for machines, some attempts have been made in this area. \citet{petrovic-matthews-2013-unsupervised} investigated transforming ordinary sentences into humorous ones, while \citet{cattle-ma-2018-recognizing} explored humor by analyzing word associations and latent semantic structures. \citet{yu-etal-2018-neural} introduced neural network models for homographic pun generation, representing an early attempt to apply fine-tuned models to humor-related tasks. Additionally, \citet{huang2022crossdialentertainingdialoguedataset} demonstrated that Crosstalk Generation remains a challenging task.

In this study, we first constructed a comprehensive dataset, \textbf{CFunSet}, by aggregating several open-source Chinese humor datasets and collecting over 20,000 jokes from Tieba—JokeBar, a section within the Chinese online forum Tieba where users share jokes and humorous content. Tieba is a large online platform where users exchange experiences and thoughts, while JokeBar specifically focuses on humor-related discussions. Subsequently, we designed appropriate humor-related tasks for the corpus. To enhance the ability of LLMs to address a wide range of humor-related tasks, we prioritized task diversity, incorporating tasks such as Humor Recognition and Evaluation, Crosstalk Generation, Joke Generation, and Joke Explanation, etc. As a result, CFunSet was compiled, comprising over 160,000 samples.

Using CFunSet, we performed supervised fine-tuning (SFT) on \textbf{Qwen2.5-7B-Instruct}, a model with strong capabilities in Chinese natural language processing, making it a suitable choice for training \textbf{CFunModel}. CFunModel demonstrated strong performance across multiple humor-related tasks. In Humor Recognition task, CFunModel achieved an accuracy of 85.98\% and for Chinese Crosstalk Response Selection, it achieved an accuracy of 91.70\% and 88.99\% for Dougen and Penggen responses, outperforming the base models. CFunModel also excelled in open-ended generation tasks, effectively generating humorous jokes and high-quality Crosstalk scripts.

The main contributions of this study are summarized as follows:
\begin{itemize}
\item We introduce \textbf{CFunSet}, a Chinese humor-related dataset comprising over 160,000 samples across multiple tasks, including Humor Recognition, Crosstalk Response Generation, Crosstalk Generation, Joke Generation, and related tasks.

\item We develop \textbf{CFunModel}, a specialized language model capable of generating and processing Chinese humor. CFunModel demonstrates superior performance compared to other large language models across multiple evaluation tasks.
\end{itemize}
\begin{table*}[ht]
\centering
\small
\setlength\tabcolsep{2pt}
\begin{tabular}{m{3cm}|m{4cm}|m{4cm}|m{4cm}} 
\toprule
\textbf{Source} &\textbf{Task} &\textbf{Input} & \textbf{Output}\\
\midrule
 Tieba-JokeBar & Joke Generation/Continuation & The first sentence of a joke. & The complete joke. \\
\midrule
 CrossDial & Crosstalk Response Selection & Crosstalk script and choices. & The correct choice.\\
 \midrule
Chumor & Humor Explanation & A joke. & Joke punchline explanation.\\
 \midrule
 HumorWB & Humor Recognition & A paragraph. & Humor label (\textbf{Humorous }or\textbf{ Not humorous}).\\
 \midrule
 Chinese-Humor-Sentiment & Crosstalk Generation & Crosstalk Script. & Continuation of the Script.\\
 \midrule
Crosstalk-Generation & Crosstalk Generation & Crosstalk script. & Continuation of the script.\\
\midrule
CCL-2019-Chinese-Humor-Computation\footnote{\url{https://github.com/DUTIR-Emotion-Group/CCL2019-Chinese-Humor-Computation}} & Joke Generation & A keyword or theme. & Joke with designated keyword or theme.\\
\bottomrule
\end{tabular}
\caption{Various tasks in CFunSet.}
\label{tab: source introduction}
\end{table*}

\section{Related Work}
\subsection{Chinese Humor Datasets}
 Humor datasets play a crucial role in advancing the investigation of humor processing in large language models (LLMs), and recent studies have made significant progress in this area. For instance, \citet{huang2022crossdialentertainingdialoguedataset} introduced \textbf{CrossDial}, a large dataset of traditional Chinese Crosstalks, which includes tasks for both Dougen and Penggen, the two primary roles in Crosstalk performances. \citet{he2024chumor20benchmarkingchinese} introduced \textbf{Chumor2.0}, a Chinese humor explanation dataset that contains jokes and their corresponding humor explanations. \citet{zeng-etal-2024-leveraging} developed \textbf{HumorWB}, a dataset collected from the Chinese social media platform Weibo, incorporating social context information such as likes and comments. \citet{li-etal-2023-language} provided scripts for Solo Crosstalk and Duet Crosstalk, while \textbf{Chinese-humor-sentiment}\footnote{\url{https://github.com/liuhuanyong/ChineseHumorSentiment}} compiled over 4,000 Duet Crosstalk scripts. However, these humor datasets have not been extensively utilized to train language models for Chinese humor processing, a gap that this study addresses.
\subsection{Humor Evaluation}
 Evaluating LLM performance on humor-related tasks is essential for assessing their ability to understand and generate humor. \citet{huang2022crossdialentertainingdialoguedataset} investigated whether LLMs can select or generate the most appropriate sentences following a given Crosstalk script. \citet{Chen_Yuan_Liu_Liu_Guan_Guo_Peng_Liu_Li_Xiao_2024} proposed two evaluation methods: one involving humor sentiment-style classification, where LLMs classify contexts as affiliative, self-enhancing, aggressive, or self-defeating; and another involving humor rewriting, where LLMs distinguish between normal and humorous responses. \citet{he2024chumor20benchmarkingchinese} introduced the task of humor comprehension, which evaluates whether LLMs can correctly explain jokes. Similarly, \citet{chen2024pretrainedlanguagemodelsunderstand} proposed four evaluation tasks: humor recognition, humor type classification, humor level classification, and punchline detection, while \citet{inoue2025laughannotationtaxonomygeneration} explored the capability of LLMs to recognize laughable contexts. However, these studies did not fully explore the potential of LLMs for humor generation and processing. In contrast, our work developed a language model capable of humor-related tasks.

\section{CFunSet Dataset}
\begin{table*}[ht]
\centering
\small
\begin{tabular}{m{3cm}|m{12cm}} 
\toprule
\textbf{Processing Target} &\textbf{Prompt}\\
\midrule
 Text filtering. & 以下是一段文本，请判断它是否具有幽默感。符合条件的文本必须是单一的一条具有幽默性的笑话，并且没有任何其他的元素。只需要输出“符合条件”或“不符合条件”。文本如下：\newline  \{文本\}\newline The following is a piece of text. Please determine whether it has a sense of humor. The text must be a single joke with humor and contain no other elements. Simply output "符合条件" (Valid) or "不符合条件" (Invalid). The text is as follows: \newline \{Text\}\\
\midrule
Theme or keyword extraction. & 用简单的词语描述概括这个笑话的主题/关键词，仅输出主题/关键词即可：\{笑话\} \newline Use simple words to describe and summarize the theme/keyword of this joke. Output only the theme/keyword: \{Joke\} \\
\bottomrule
\end{tabular}
\caption{Prompt template for data processing.}
\label{tab: prompt}
\end{table*}

\subsection{Overview}
We meticulously selected several open-source datasets as the foundation of \textbf{CFunSet}. Additionally, data samples were collected from Tieba—JokeBar, a Chinese online forum where a variety of jokes are shared. The data sources and tasks in CFunSet are summarized in Table \ref{tab: source introduction}.
The dataset creation process consists of two primary stages:
 \begin{itemize}
 \item \textbf{Data Collection and Processing} \
Several existing Chinese humor-related datasets were collected and refined. Data samples were extracted from Tieba—JokeBar and subsequently cleaned and processed to ensure consistency and usability.
\item \textbf{Humor-Related Task Design} \
 To enhance the model's humor-related capacity, a diverse set of tasks was designed for each type of data sample.
 \end{itemize}

\subsection{Data Collection and Processing}
\label{3.2}
We first cleared up datasets including CrossDial\citep{huang2022crossdialentertainingdialoguedataset}, Chumor2.0\citep{he2024chumor20benchmarkingchinese}, HumorWB\citep{zeng-etal-2024-leveraging}, Chinese-Humor-Sentiment\citep{chinesehumorsentiment}, Crosstalk-Generation\citep{li-etal-2023-language} and CCL2019-Chinese-Humor-Computation\citep{ccl2019}. We then collected posts and comments from Tieba—JokeBar that contained more than 50 characters. To ensure data quality, Qwen2.5-7B-Instruct was employed to filter out non-humorous content, retaining only texts classified as jokes. For data obtained from CCL2019-Chinese-Humor-Computation, Qwen2.5-7B-Instruct was also utilized to extract keywords and themes, which facilitated the subsequent task design stage, as discussed in Section \ref{3.3}.

The detailed data processing prompts are presented in Table \ref{tab: prompt}.

\begin{table*}[ht]
\centering
\small
\setlength\tabcolsep{4pt}
\renewcommand{\arraystretch}{1.1}
\begin{tabular}{>{\centering\arraybackslash}p{3cm}|
                >{\centering\arraybackslash}p{3.5cm}|
                >{\centering\arraybackslash}p{3.5cm}|
                >{\centering\arraybackslash}p{3.5cm}} 
\toprule
\textbf{Model} & \textbf{CrossDial-Dougen(ACC)} & \textbf{CrossDial-Penggen(ACC)} &\textbf{HumorWB(ACC)}\\
\midrule
GPT-4o & 79.67 & 73.88 & 83.41  \\
GPT-4o mini & 74.14 & 67.45 & 84.78 \\
DeepSeek-V3  & 83.66 & 78.16 & 85.15 \\
\midrule
ERNIE  & 84.54 & - & - \\
RoBERTa  & - & 76.19 & - \\
\midrule
Qwen2.5-7B-Instruct & 24.74 & 20.87 & 79.56 \\
CFunModel(ours) & \textbf{91.70} & \textbf{88.99} & \textbf{85.98} \\
\bottomrule
\end{tabular}
\caption{Model performance on Crosstalk Response Selection and Humor Recognition tasks. ERNIE and RobERTa is the best models \citet{huang2022crossdialentertainingdialoguedataset} had trained for Dougen and Penggen Response Generation.}
\label{tab: result}
\end{table*}
\subsection{Humor-related Task Design}
\label{3.3}
Our humor-related tasks were designed to ensure a high degree of alignment with the data and a diverse range of task formats. Each sample in \textbf{CFunSet} consists of three components:
\begin{itemize}
\item \textbf{Instruction}: A description and prompt specifying the task type.
\item \textbf{Input}: Relevant data upon which the output should be based.
\item \textbf{Output}: The expected response generated for the task.
\end{itemize}
 Examples of data samples and their structures are shown in Appendix(Table \ref{tab: CFunset Examples.}). And the following section provides a detailed description of the data and task design.

\paragraph{Tieba—JokeBar}
 Jokes containing more than 50 characters were selected, with the first sentence extracted as the Input. The task assigned to the model is \textbf{Joke Generation/Continuation}, where the model should generate or continue the joke based on the provided input.
\paragraph{CrossDial}
 CrossDial comprises source texts, multiple-choice options, and target texts. The task is \textbf{Crosstalk Response Selection}, where the model is tasked with selecting the most appropriate response from the options as the response for the given Crosstalk script segment.
 
\paragraph{Chumor2.0}
 Chumor2.0 contains jokes and their corresponding punchline explanations. The jokes and explanations are used as the input and output for the \textbf{Humor Explanation} task, where the model is expected to provide accurate explanations for the jokes.
 
\paragraph{HumorWB}
HumorWB comprises texts sourced from the Chinese platform Weibo, along with labels indicating whether the texts are humorous. The texts from HumorWB are used as the input in the \textbf{Humor Recognition} task, where the model is required to determine whether a given text is humorous.

\paragraph{Chinese-Humor-Sentiment/Crosstalk-Generation}
Both datasets consist of Crosstalk scripts, and we specifically selected Dual Crosstalk scripts, which capture the dialogue between the two roles, Dougen and Penggen. The task is \textbf{Crosstalk Generation/Continuation}, where the model is tasked with completing a Crosstalk script based on a given script segment.

\paragraph{CCL2019-Chinese-Humor-Computation}
Human-written jokes from Task 1 and jokes labeled as 5 (the highest level of humor) from Task 2 of CCL2019-Chinese-Humor-Computation were selected as the input. Qwen2.5-7B-Instruct was utilized to extract keywords or themes, and the corresponding task is \textbf{Joke Generation}, where the model is required to generate a joke based on the extracted keyword or theme.
\begin{table*}[ht]
\centering
\small
\setlength\tabcolsep{2pt}
\begin{tabular}{m{3cm}|m{6cm}|m{6cm}} 
\toprule
\textbf{Prompt} & \textbf{CFunModel} & \textbf{Qwen2.5-7B-Instruct} \\
\midrule
生成一个主题为家庭琐事的笑话。\newline \newline Generate a joke on the theme of family trifles. &  儿子：“妈妈，我想你，因为我们好几天没见面了。” \newline 妈妈：“孩子，我也好想你，因为已经一个星期没做家务了。”\newline Son: "Mom, I miss you because we haven't seen each other for days."\newline Mom: "Sweetie, I miss you too because I haven't done any housework for a week." & 为什么家里的垃圾筒总是不满？因为它每天都在“家庭琐事”的填满中，忙得不亦乐乎！\newline Why is the trash can at home always full? Because it’s busy being filled with "family trifles" every day, having the time of its life!\\
\midrule
生成一个主题为风水的笑话。\newline  Generate a joke on the theme of Fengshui.\newline (Feng Shui is an ancient Chinese practice that aims to harmonize people with their environment by arranging objects to promote positive energy.) & 我有一个朋友，他非常相信风水。 有一次他家房子装修完毕后，他请风水大师来看一看。 风水大师说：“你家房子的方位很不吉利，需要重新布置一下。” 我朋友问：“怎么布置？” 风水大师说：“你把床放在衣柜的上面。” \newline  I have a friend who is very superstitious about Feng Shui. One time, after his house was renovated, he invited a Feng Shui master to take a look. The Feng Shui master said: "The position of your house is very unlucky. You need to rearrange it." My friend asked: "How should I rearrange it?" The Feng Shui master said: "Put your bed on top of the wardrobe."
 & 风水大师来到客户家，客户指着墙上的一幅画说：“风水大师，这幅画风水好吗？”\newline 风水大师看了一眼，认真地说：“不好。” \newline 客户疑惑地问：“为什么呢？”\newline 风水大师回答：“因为这是一幅画，而你把它挂在墙上，说明你不懂风水！”\newline  A feng shui master visited a client’s home. The client pointed to a painting on the wall and asked, “Master, does this painting have good feng shui?”\newline The feng shui master took a look and replied seriously, “No, it doesn’t.” \newline The client asked in confusion, “Why not?” \newline The feng shui master answered, “Because it’s just a painting, and the fact that you hung it on the wall shows that you don’t understand feng shui!”\\
\bottomrule
\end{tabular}
\caption{Joke examples generated by CFunModel and QWen2.5-7B-Instruct. The humor in the example on family trifles generated by CFunModel lies in the contrast between the son's sincere expression of missing his mom and the mom's playful response, implying that her longing is more about needing help with housework. The example on superstition employs an absurd piece of advice from a so-called "master" to highlight its irrationality. In contrast, the examples generated by Qwen2.5-7B-Instruct lack humor, merely describing ordinary events.}
\label{tab: joke example}
\end{table*}

\section{CFunModel}
\textbf{CFunModel} is trained to address a variety of humor-related tasks, including Crosstalk Response Selection, Humor Recognition, Crosstalk and Joke Generation. The prompt structure is a combination of \textbf{Instruction} and \textbf{Input}, as described in Section \ref{3.2}.

The base model used is \texttt{Qwen2.5-7B-Instruct}. Supervised fine-tuning was performed using \textbf{CFunSet} as the training data for \texttt{2} epochs, with a learning rate of \texttt{2e-5} and a batch size of \texttt{128}. The training process was conducted on four A40 GPUs, and we set the temperature to 0 during the evaluation stage.

\section{Experiment}
\subsection{Setup}

For baseline comparisons, the models selected include \textbf{GPT-4o}\footnote{\url{https://openai.com/index/gpt-4o-system-card/}}, \textbf{GPT-4o Mini}\footnote{\url{https://openai.com/index/gpt-4o-mini-advancing-costefficient-intelligence/}}, and \textbf{DeepSeek-V3}\citep{deepseekai2025deepseekv3technicalreport}. The temperature was set to 1 during the evaluation stage.

Several tasks were selected to evaluate the performance of the models:
\begin{itemize}
\item \textbf{Humor Recognition} \\
Since the original HumorWB dataset did not have predefined training and test splits, I manually divided it into 4,200 samples for training and 1,091 samples for testing. The model is tasked with determining whether the given texts are humorous. \textbf{Accuracy} is used to assess the model's performance on this task.

\item \textbf{Crosstalk Response Selection} \\  
Test data from CrossDial was used, comprising 4,664 samples for Dougen response selection and 2,688 samples for Penggen response selection. \textbf{Accuracy} is calculated to evaluate the model’s effectiveness in selecting the most appropriate response.  

\item \textbf{Joke Generation and Continuation} \\  
Since Joke Generation and Continuation are open-ended tasks, traditional metrics such as \textbf{BLEU}\citep{BLEU} are insufficient for accurately assessing the model's performance. Therefore, case study is performed to qualitatively evaluate the model's ability in these tasks.  

\item \textbf{Crosstalk Generation and Continuation} \\  
Similar to Joke Generation and Continuation, case study is employed to qualitatively evaluate the model's capability in Crosstalk Generation and Continuation tasks.  
\end{itemize}

\begin{table*}[ht]
\centering
\small
\begin{tabular}{m{8cm}|m{8cm}} 
\toprule
\textbf{Crosstalk Generated by CFunModel} & \textbf{Translation} \\
\midrule
 甲：我今天跟你说，足球，这个东西，你别看它那么小，但是，它里面可有大学问啊！\newline 乙：你先说说，足球里面都有什么学问？\newline 甲：首先，你要知道，足球是圆的，它象征着地球，这说明我们地球人要像足球一样，走南闯北，四海为家，这样，我们地球人才能更团结。 \newline 乙：这个我知道。 \newline 甲：然后，你要知道，足球是用牛皮做的，这说明我们地球人要像牛一样，任劳任怨，无私奉献，这样，我们地球人才能更和谐。 \newline 乙：这个我更知道。 \newline 甲：最后，你要知道，足球比赛时，裁判员会吹哨子，这说明我们地球人要像哨子一样，公平公正，一视同仁，这样，我们地球人才能更正义。 \newline 乙：这个我最知道。 \newline 甲：怎么样，足球里面有学问吧！ \newline 乙：嗯，有，有，有。 & A: Let me tell you something today. Football may seem small, but there’s a lot of knowledge inside it!\newline B: Oh? What kind of knowledge is in football?\newline A: First, you should know that a football is round, symbolizing the Earth. This means that we, as Earthlings, should be like a football—traveling around, feeling at home wherever we go. This way, we Earthlings can be more united.\newline B: I know that.\newline A: Then, you should know that football is made of cowhide, which implies that we, as Earthlings, should be like cows—hardworking, selfless, and dedicated. This way, we Earthlings can be more harmonious.\newline B: I know that even better.\newline A: Lastly, during a football match, the referee blows the whistle, which shows that we, as Earthlings, should be like a whistle—fair, just, and treating everyone equally. This way, we Earthlings can be more righteous.\newline B: I know that best.\newline A: So, what do you think? Football contains a lot of knowledge, right?\newline B: Hmm... yes, yes, it does.\\
\bottomrule
\end{tabular}
\caption{Crosstalk example generated by CFunModel. The prompt is "\textit{Generate a Crosstalk with the theme of football} (生成一段主题为足球的相声).}
\label{tab: cross example}
\end{table*}

\subsection{Results}
Overall, CFunModel outperforms the baseline models across multiple tasks, as presented in Table \ref{tab: result}. In the Crosstalk Response Generation task, CFunModel achieves an accuracy of 91.70\% for Dougen responses and 88.99\% for Penggen responses, exceeding the performance of GPT-4o, GPT-4o Mini, and DeepSeek-V3, thereby demonstrating its superior capability in generating contextually appropriate and humorous responses. Compared to the base model Qwen2.5-7B-Instruct, CFunModel exhibits significant improvement, indicating that the training process effectively enhances the base model’s ability to handle humor-related tasks.

In the Humor Recognition task, CFunModel achieves a notable accuracy of 85.98\%, surpassing the performance of GPT-4o, GPT-4o Mini, and DeepSeek-V3. These results underscore the robustness of CFunModel in handling a variety of humor-related tasks.

\subsection{Case Study}
 To further evaluate the ability of CFunModel in Crosstalk and Joke Generation, several Crosstalk scripts and jokes were generated and compared with those produced by Qwen2.5-7B-Instruct. Both models were prompted to generate jokes on the topics of household affairs and superstition, as shown in Table \ref{tab: joke example}. The results demonstrate that the jokes generated by CFunModel exhibit greater coherence and humor. However, the jokes generated by Qwen2.5-7B-Instruct often describe scenarios that are somewhat implausible and fail to convey humor effectively. These findings underscore the importance of training with CFunSet in enhancing the base model’s performance on humor-related tasks.
 
In Crosstalk-related tasks, CFunModel also demonstrates superior performance. Examples of Crosstalk Generation are provided in Table \ref{tab: cross example}. The results indicate that the examples generated by CFunModel construct engaging conversations between Dougen and Penggen, where Dougen effectively introduces amusing topics and creates humorous situations and Penggen asks relevant questions, responds appropriately, and enhances the humor of the conversation.

\section{Conclusion}
This paper introduces \textbf{CFunSet}, the first Chinese humor-related multi-task dataset, which encompasses a diverse range of tasks, including humor recognition, Crosstalk response selection, joke generation, joke explanation, etc. Using CFunSet, we developed \textbf{CFunModel}, the first open-source language model capable of generating and processing Chinese humor. The model demonstrates strong performance across various humor-related challenging tasks. CFunModel surpasses existing models, offering enhanced functionality and applicability in Chinese humor.

\section*{Limitations}
Significant efforts were made to clean the data collected from Tieba-JokeBar; however, some instances of misspellings, grammatical errors, and incorrect punctuation may still be present in the final dataset.

\section*{Ethical Considerations}
This study introduces CFunSet. Regarding intellectual property, all data sources used in this dataset are publicly available on the internet.

Although the dataset has been reviewed and multiple techniques have been employed to eliminate inappropriate language, there remains the possibility that some content may inadvertently cause offense.

\bibliography{custom}

\begin{thebibliography}{14}
\providecommand{\natexlab}[1]{#1}

\bibitem[{Cattle and Ma(2018)}]{cattle-ma-2018-recognizing}
Andrew Cattle and Xiaojuan Ma. 2018.
\newblock \href {https://aclanthology.org/C18-1157/} {Recognizing humour using word associations and humour anchor extraction}.
\newblock In \emph{Proceedings of the 27th International Conference on Computational Linguistics}, pages 1849--1858, Santa Fe, New Mexico, USA. Association for Computational Linguistics.

\bibitem[{Chen et~al.(2024{\natexlab{a}})Chen, Li, Liang, Xiao, Liu, and Chen}]{chen2024pretrainedlanguagemodelsunderstand}
Yuyan Chen, Zhixu Li, Jiaqing Liang, Yanghua Xiao, Bang Liu, and Yunwen Chen. 2024{\natexlab{a}}.
\newblock \href {https://arxiv.org/abs/2407.04105} {Can pre-trained language models understand chinese humor?}
\newblock \emph{Preprint}, arXiv:2407.04105.

\bibitem[{Chen et~al.(2024{\natexlab{b}})Chen, Yuan, Liu, Liu, Guan, Guo, Peng, Liu, Li, and Xiao}]{Chen_Yuan_Liu_Liu_Guan_Guo_Peng_Liu_Li_Xiao_2024}
Yuyan Chen, Yichen Yuan, Panjun Liu, Dayiheng Liu, Qinghao Guan, Mengfei Guo, Haiming Peng, Bang Liu, Zhixu Li, and Yanghua Xiao. 2024{\natexlab{b}}.
\newblock \href {https://doi.org/10.1609/aaai.v38i16.29736} {Talk funny! a large-scale humor response dataset with chain-of-humor interpretation}.
\newblock \emph{Proceedings of the AAAI Conference on Artificial Intelligence}, 38(16):17826--17834.

\bibitem[{DeepSeek-AI et~al.(2025)DeepSeek-AI, Liu, Feng, Xue, Wang, Wu, Lu, Zhao, Deng, Zhang, Ruan, Dai, Guo, Yang, Chen, Ji, Li, Lin, Dai, Luo, Hao, Chen, Li, Zhang, Bao, Xu, Wang, Zhang, Ding, Xin, Gao, Li, Qu, Cai, Liang, Guo, Ni, Li, Wang, Chen, Chen, Yuan, Qiu, Li, Song, Dong, Hu, Gao, Guan, Huang, Yu, Wang, Zhang, Xu, Xia, Zhao, Wang, Zhang, Li, Wang, Zhang, Zhang, Tang, Li, Tian, Huang, Wang, Zhang, Wang, Zhu, Chen, Du, Chen, Jin, Ge, Zhang, Pan, Wang, Xu, Zhang, Chen, Li, Lu, Zhou, Chen, Wu, Ye, Ye, Ma, Wang, Zhou, Yu, Zhou, Pan, Wang, Yun, Pei, Sun, Xiao, Zeng, Zhao, An, Liu, Liang, Gao, Yu, Zhang, Li, Jin, Wang, Bi, Liu, Wang, Shen, Chen, Zhang, Chen, Nie, Sun, Wang, Cheng, Liu, Xie, Liu, Yu, Song, Shan, Zhou, Yang, Li, Su, Lin, Li, Wang, Wei, Zhu, Zhang, Xu, Xu, Huang, Li, Zhao, Sun, Li, Wang, Yu, Zheng, Zhang, Shi, Xiong, He, Tang, Piao, Wang, Tan, Ma, Liu, Guo, Wu, Ou, Zhu, Wang, Gong, Zou, He, Zha, Xiong, Ma, Yan, Luo, You, Liu, Zhou, Wu, Ren, Ren, Sha, Fu, Xu, Huang, Zhang, Xie, Zhang, Hao,
  Gou, Ma, Yan, Shao, Xu, Wu, Zhang, Li, Gu, Zhu, Liu, Li, Xie, Song, Gao, and Pan}]{deepseekai2025deepseekv3technicalreport}
DeepSeek-AI, Aixin Liu, Bei Feng, Bing Xue, Bingxuan Wang, Bochao Wu, Chengda Lu, Chenggang Zhao, Chengqi Deng, Chenyu Zhang, Chong Ruan, Damai Dai, Daya Guo, Dejian Yang, Deli Chen, Dongjie Ji, Erhang Li, Fangyun Lin, Fucong Dai, and 181 others. 2025.
\newblock \href {https://arxiv.org/abs/2412.19437} {Deepseek-v3 technical report}.
\newblock \emph{Preprint}, arXiv:2412.19437.

\bibitem[{DUTIR-Emotion-Group(2019)}]{ccl2019}
DUTIR-Emotion-Group. 2019.
\newblock \href {https://github.com/DUTIR-Emotion-Group/CCL2019-Chinese-Humor-Computation} {Ccl2019-chinese-humor-computation}.

\bibitem[{He et~al.(2024)He, He, Bai, Liu, Sun, Tang, Wang, Xia, Mihalcea, and Deng}]{he2024chumor20benchmarkingchinese}
Ruiqi He, Yushu He, Longju Bai, Jiarui Liu, Zhenjie Sun, Zenghao Tang, He~Wang, Hanchen Xia, Rada Mihalcea, and Naihao Deng. 2024.
\newblock \href {https://arxiv.org/abs/2412.17729} {Chumor 2.0: Towards benchmarking chinese humor understanding}.
\newblock \emph{Preprint}, arXiv:2412.17729.

\bibitem[{Huang et~al.(2022)Huang, Du, and Wan}]{huang2022crossdialentertainingdialoguedataset}
Baizhou Huang, Shikang Du, and Xiaojun Wan. 2022.
\newblock \href {https://arxiv.org/abs/2209.01370} {Crossdial: An entertaining dialogue dataset of chinese crosstalk}.
\newblock \emph{Preprint}, arXiv:2209.01370.

\bibitem[{Inoue et~al.(2025)Inoue, Elmers, Lala, and Kawahara}]{inoue2025laughannotationtaxonomygeneration}
Koji Inoue, Mikey Elmers, Divesh Lala, and Tatsuya Kawahara. 2025.
\newblock \href {https://arxiv.org/abs/2501.16635} {Why do we laugh? annotation and taxonomy generation for laughable contexts in spontaneous text conversation}.
\newblock \emph{Preprint}, arXiv:2501.16635.

\bibitem[{Li et~al.(2023)Li, Wu, Liu, Xie, Tiwari, and Wang}]{li-etal-2023-language}
Jianquan Li, XiangBo Wu, Xiaokang Liu, Qianqian Xie, Prayag Tiwari, and Benyou Wang. 2023.
\newblock \href {https://doi.org/10.18653/v1/2023.acl-long.419} {Can language models make fun? a case study in {C}hinese comical crosstalk}.
\newblock In \emph{Proceedings of the 61st Annual Meeting of the Association for Computational Linguistics (Volume 1: Long Papers)}, pages 7581--7596, Toronto, Canada. Association for Computational Linguistics.

\bibitem[{liuhuanyong(2018)}]{chinesehumorsentiment}
liuhuanyong. 2018.
\newblock \href {https://github.com/liuhuanyong/ChineseHumorSentiment} {Chinesehumorsentiment}.

\bibitem[{Papineni et~al.(2002)Papineni, Roukos, Ward, and Zhu}]{BLEU}
Kishore Papineni, Salim Roukos, Todd Ward, and Wei-Jing Zhu. 2002.
\newblock \href {https://doi.org/10.3115/1073083.1073135} {Bleu: a method for automatic evaluation of machine translation}.
\newblock In \emph{Proceedings of the 40th Annual Meeting on Association for Computational Linguistics}, ACL '02, page 311–318, USA. Association for Computational Linguistics.

\bibitem[{Petrovi{\'c} and Matthews(2013)}]{petrovic-matthews-2013-unsupervised}
Sa{\v{s}}a Petrovi{\'c} and David Matthews. 2013.
\newblock \href {https://aclanthology.org/P13-2041/} {Unsupervised joke generation from big data}.
\newblock In \emph{Proceedings of the 51st Annual Meeting of the Association for Computational Linguistics (Volume 2: Short Papers)}, pages 228--232, Sofia, Bulgaria. Association for Computational Linguistics.

\bibitem[{Yu et~al.(2018)Yu, Tan, and Wan}]{yu-etal-2018-neural}
Zhiwei Yu, Jiwei Tan, and Xiaojun Wan. 2018.
\newblock \href {https://doi.org/10.18653/v1/P18-1153} {A neural approach to pun generation}.
\newblock In \emph{Proceedings of the 56th Annual Meeting of the Association for Computational Linguistics (Volume 1: Long Papers)}, pages 1650--1660, Melbourne, Australia. Association for Computational Linguistics.

\bibitem[{Zeng et~al.(2024)Zeng, Li, Yang, and Lin}]{zeng-etal-2024-leveraging}
Zeyuan Zeng, Zefeng Li, Liang Yang, and Hongfei Lin. 2024.
\newblock \href {https://aclanthology.org/2024.lrec-main.908/} {Leveraging social context for humor recognition and sense of humor evaluation in social media with a new {C}hinese humor corpus - {H}umor{WB}}.
\newblock In \emph{Proceedings of the 2024 Joint International Conference on Computational Linguistics, Language Resources and Evaluation (LREC-COLING 2024)}, pages 10393--10402, Torino, Italia. ELRA and ICCL.

\end{thebibliography}
\appendix

\section*{Appendix}
\label{sec:appendix}

\centering
\small
\onecolumn
\begin{longtable}{m{2cm}|m{13cm}} 
\hline
\textbf{Source} & \textbf{Example} \\
\hline
\endfirsthead
\hline
\textbf{Source} & \textbf{Example} \\
\hline
\endhead
\hline
 Tieba-JokeBar & \textbf{Instruction:}\newline 我将给你笑话的第一句话，请你生成整个笑话。笑话的第一句话如下：I will give you the first sentence of a joke, and you need to generate the whole joke. Here’s the first sentence:\newline \textbf{Input:} \newline 我对女儿说：再看10分钟的动画片就该睡觉了。\newline I told my daughter: "You can watch the cartoon for 10 more minutes, and then it's bedtime."\newline \textbf{Output:}\newline 她抗议说时间太短。“那就600秒，够长了吧？”我说。女儿说：“够了，够了。” \newline She protested, saying the time was too short. "How about 600 seconds? That's a lot longer, right?" I said. My daughter quickly replied, "Oh, that’s long enough!" \\
\midrule
CrossDial& \textbf{Instruction:}\newline给定一段相声台词，请从多个备选项中选择最合适的逗哏回复。 Given a crosstalk script, choose the most suitable Dougen reply from multiple options.)\newline \textbf{Input:} \newline 已有台词：我是肖飞。|噢。|给大家说一段。|好。|吃五谷杂粮没有不得病的。|是。|我是大病上医院，小病上药店。|小病干嘛不找大夫呀？|挂号费够买药的了。|噢。\newline候选回复：早上天还没亮，三狗屁就来到了工程队。\newline 反正很重，去买药，下台阶儿。\newline 那天我脚气犯了，上药店买点儿药吧。\newline 都快半夜了，上哪找药店去？ \newline Existing Script: I am Xiao Fei.|Oh.|Let me tell you a story.|Okay.|Eating grains and cereals, everyone gets sick sometimes.|Yes.|When I have a serious illness, I go to the hospital. For minor illnesses, I go to the pharmacy.|Why not see a doctor for minor illnesses?|The registration fee is enough to buy medicine.|Oh.\newline Candidate Replies: Early in the morning before dawn, San Gou Pi went to the construction team.\newline It's heavy anyway, go buy medicine, and watch your step.\newline That day, I had athlete's foot, so I went to the pharmacy to get some medicine.\newline It’s almost midnight, where can I find a pharmacy?\newline \textbf{Output:}\newline那天我脚气犯了，上药店买点儿药吧。\newline That day, I had athlete's foot, so I went to the pharmacy to get some medicine.\\
 \midrule
Chumor& \textbf{Instruction:}\newline请阅读以下文字，分析其幽默的原因.幽默性指该文本是否可能引起读者发笑，或通过语言技巧（如双关语、讽刺、夸张、荒诞或逻辑上的意外等等方式）营造幽默效果。请你写出以下文字幽默的原因：Please read the following text and analyze the reason for its humor. Humor refers to whether the text is likely to make the reader laugh or create a humorous effect through linguistic techniques (such as puns, sarcasm, exaggeration, absurdity, or unexpected logic, etc.). Please explain the reason for the humor in the following text:\newline \textbf{Input:} \newline 比起段子，应该还是我的论文更加搞笑一点。\newline Compared to jokes, my thesis is probably even funnier.\newline \textbf{Output:}\newline 这个笑话的幽默之处在于自嘲和反差。作者通过自嘲自己的论文质量不高，反而比段子更搞笑，制造了意想不到的反差效果。\newline The humor in this joke lies in self-deprecation and contrast. The author humorously mocks the quality of their own thesis by suggesting that it is even funnier than a joke, creating an unexpected contrast that generates humor.\\
 \midrule
HumorWB & \textbf{Instruction:}\newline以下是一段文本，请分析它是否具有幽默性。幽默性指该文本是否可能引起读者发笑，或通过语言技巧（如双关语、讽刺、夸张、荒诞或逻辑上的意外）营造幽默效果。只需要输出“幽默”或“不幽默”。The following is a piece of text. Please analyze whether it has humor. Humor refers to whether the text is likely to make the reader laugh or create a humorous effect through linguistic techniques (such as puns, sarcasm, exaggeration, absurdity, or unexpected logic, etc.). Only output "Humorous" or "Not humorous". \newline \textbf{Input:}\newline问一下已经步入社会的大人们，假如有机会带着现在的记忆回到高一那年，你最想做什么？\newline Ask the adults who have already entered society: If you had the chance to go back to your first year of high school with your current memories, what would you most want to do?\newline \textbf{Output:}\newline不幽默\newline Not humorous\\
 \midrule
Chinese-Humor-Sentiment and Crosstalk-Generation & \textbf{Instruction:}\newline以下是一段相声的前一半内容，请你根据给定的部分续写完成这段相声。\newline The following is the first half of a crosstalk script. Please continue and complete the script based on the given part.\newline \textbf{Input:} \newline0:“常言道，烟酒不分家。来，抽一根儿!”\newline 1:“你小子真鬼，用你的孬烟换我的好酒!”\newline 0:“吝啬鬼，你能喝什么好酒?”\newline :“谁不知道你呀，犯了瘾十烟屁股吸。”\newline 0:“你馋的不要脸，蹭人家的酒喝。”\newline  0: As the saying goes, smoking and drinking go hand in hand. Come on, have a puff!\newline 1: You’re so sneaky, trading your lousy cigarettes for my good wine!\newline 0: Stingy guy, what good wine can you even afford?\newline 1: Everyone knows you! When you’re desperate, you even smoke cigarette butts.\newline 0: And you’re shamelessly greedy, always mooching off others for a drink.\newline \textbf{Output:}\newline1:“你讨厌讨厌真讨厌，公共场合放毒烟;害人害己害环境，吞云吐雾造污染。浪费开支烧金钱，口干舌燥得咽炎;慢性自杀损寿命，戒烟戒烟快戒烟!”\newline 0:“你混蛋混蛋真混蛋，一喝就高吐一滩;伤头伤胃伤心肝，醉生梦死早完蛋……”\newline 1:“你完蛋!”\newline 0:“你完蛋!”\newline 1:“你是烟鬼，你滚蛋!”\newline 0:“你是酒鬼，你滚蛋!”\newline 1: You’re annoying, really annoying! Smoking in public spreads toxic fumes; harming others, yourself, and the environment. Blowing clouds pollutes the air, burning money and getting throat infections. It's slow suicide, shortening your life—quit smoking, quit smoking now!\newline 0: You’re a real pain, a real pain! Every time you drink, you puke everywhere. Hurts your head, stomach, and liver—living a drunken life will ruin you sooner or later…\newline 1: You’re done for!\newline 0: You’re done for!\newline 1: You’re a smoking addict, get lost!\newline 0: You’re a drunkard, you get lost!\\
 \midrule
CCL-2019-Chinese-Humor-Computation Task 1 & \textbf{Instruction:}\newline生成一个关键词为“砍价”的笑话。\newline Generate a joke with the keyword "bargaining." \newline \textbf{Input:} \newline -\newline \textbf{Output:}\newline老板：一块五一斤。\newline 我：太贵了，五块钱三斤吧。\newline 老板：不行不行。\newline Shopkeeper: One and a half yuan per pound.\newline Me: That's too expensive. How about five yuan for three pounds?\newline Shopkeeper: No way, no way.\\
 \midrule
CCL-2019-Chinese-Humor-Computation Task 2 & \textbf{Instruction:}\newline生成一个主题为偷窃的笑话。\newline Generate a joke with the theme "theft." \newline \textbf{Input:} \newline -\newline \textbf{Output:}\newline小偷偷了只鸡，在河边拔毛。 有人经过，小偷忙将鸡扔进河，并向来人解释：这只鸡在游泳，我帮它看衣服。\newline A thief stole a chicken and was plucking its feathers by the river. When someone passed by, the thief quickly threw the chicken into the river and explained to the passerby: "The chicken is swimming, and I’m just watching its clothes for it."\\
\bottomrule
\caption{Examples of CFunSet.}
\label{tab: CFunset Examples.}
\end{longtable}

\end{CJK}
\end{document}